\documentclass[accepted]{uai2023} 

\usepackage[american]{babel}

\usepackage{natbib} 
\usepackage{mathtools} 
\usepackage{booktabs} 
\usepackage{tikz} 

\usepackage{amsmath,amsfonts,bm}

\def\eqref#1{equation~\ref{#1}}

\def\1{\bm{1}}

\DeclareMathAlphabet{\mathsfit}{\encodingdefault}{\sfdefault}{m}{sl}
\SetMathAlphabet{\mathsfit}{bold}{\encodingdefault}{\sfdefault}{bx}{n}

\DeclareMathOperator*{\argmax}{arg\,max}

\usepackage{hyperref}
\usepackage{url}

\usepackage{amsmath,amssymb,amsfonts,mathrsfs,amsthm,bbm}
\usepackage{algorithm}
\usepackage{algorithmic}
\usepackage{graphicx}
\usepackage{textcomp}
\usepackage{xcolor}
\usepackage{paralist}

\usepackage{subfigure}
\usepackage{booktabs}  
\usepackage{adjustbox}
\usepackage{caption}
\usepackage{amsmath}
\newcommand\numberthis{\addtocounter{equation}{1}\tag{\theequation}}
\usepackage{xcolor}
\usepackage{enumitem}
\usepackage[normalem]{ulem}

\newtheorem{theorem}{Theorem}
\newtheorem{definition}{Definition}

\title{Using Semantic Information for Defining and Detecting OOD Inputs}
\author[1]{Ramneet Kaur}
\author[1]{Xiayan Ji*}
\author[1]{Souradeep Dutta*}
\author[1]{Michele Caprio*}
\author[1]{Yahan Yang}
\author[2]{Elena Bernardis}
\author[1]{\\ Oleg Sokolsky}
\author[1]{Insup Lee}

\affil[1]{%
    PRECISE Lab\\
    Computer and Information Science Dept.\\
    University of Pennsylvania\\
    Philadelphia, Pennsylvania, USA
}
\affil[2]{%
    Dermatology Dept.\\
    University of Pennsylvania\\
    Philadelphia, Pennsylvania, USA
}
  
\begin{document}
\maketitle
\def\thefootnote{*}\footnotetext{These authors contributed equally to this work.}\def\thefootnote{\arabic{footnote}}
\begin{abstract}

As machine learning models continue to achieve impressive performance across different tasks, the importance of effective anomaly detection for such models has increased as well. 
It is common knowledge that even well-trained models lose their ability to function effectively on out-of-distribution inputs. Thus, out-of-distribution (OOD) detection has received some attention recently. In the vast majority of cases, it uses the distribution estimated by the training dataset for OOD detection. We demonstrate that the current detectors inherit the biases in the training dataset, unfortunately. This is a serious impediment, and can potentially restrict the utility of the trained model. 
This can render the current OOD detectors impermeable to inputs lying outside the training distribution but with the same semantic information (e.g. training class labels).\
To remedy this situation, we begin by defining what should ideally be treated as an OOD, by connecting inputs with their semantic information content.  We perform OOD detection on semantic information extracted from the training data of MNIST and COCO datasets and show that it not only reduces false alarms but also significantly improves the detection of OOD inputs with spurious features from the training data. 
\end{abstract}

\section{Introduction}
\label{sec:intro}
Machine learning models have achieved remarkable success in accomplishing different tasks across modalities such as image classification~\citep{img-classification}, speech recognition~\citep{speech-recog}, and natural language processing~\citep{text-analysis}. It is however known, that these  models are unreliable on samples that are less likely to occur, according to the model's \textit{in-distribution} estimated from its training data~\citep{baseline}. Detection of these \textit{out-of-distribution (OOD)} inputs is important for the deployment of machine learning models in safety-critical domains such as autonomous driving~\citep{ML-App7}, and medical diagnosis~\citep{NNclinically}. OOD detection has, therefore, gained a lot of attention recently~\citep{codit,odin,mahala,aux,kaur2021all}. 

\begin{figure*}
    \centering
    \includegraphics[width=0.7\textwidth]{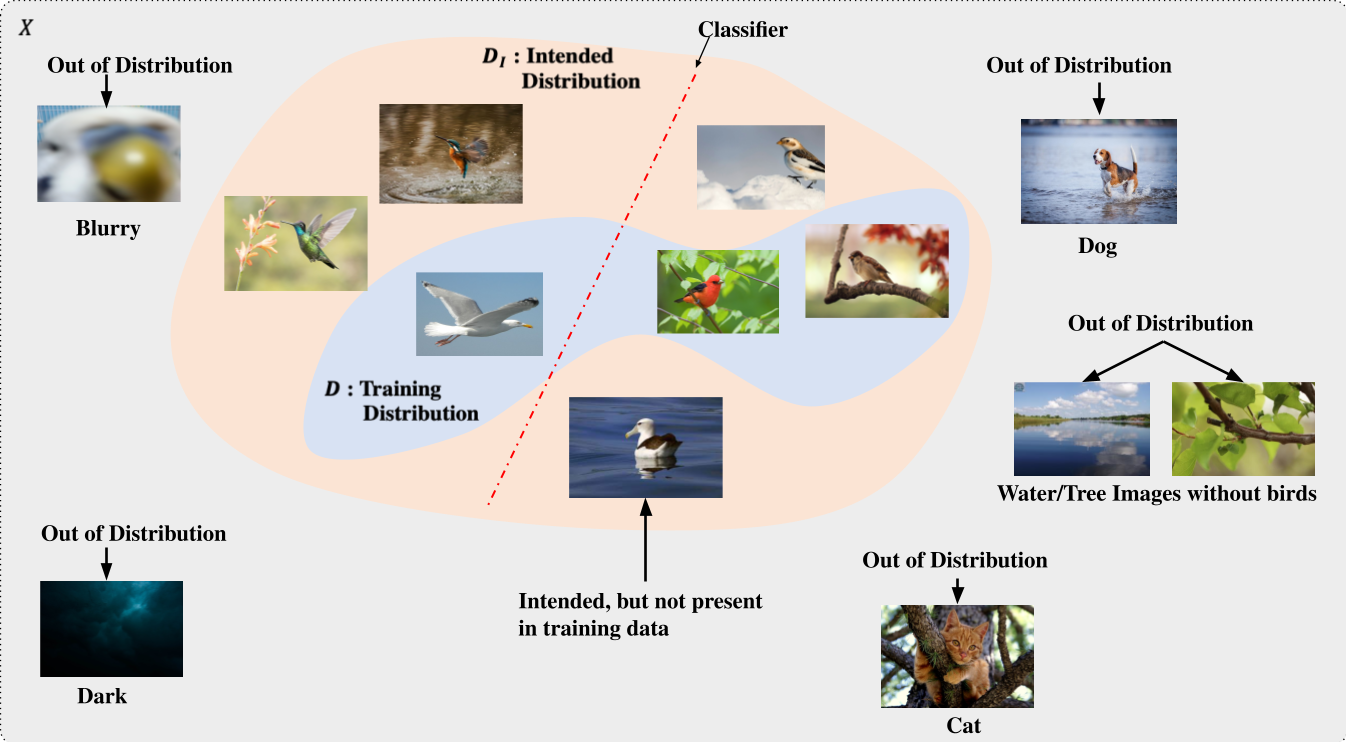}
    \caption{The intended distribution has a much higher variability in terms of the samples it covers, when compared to the training distribution. The classifier trained to classify birds in \{sitting birds, flying birds\} is expected to generalize well for the intended distribution, which has birds sitting on trees, snow or water. OOD inputs are the ones which are unlikely to occur from the point of the intended distribution $\mathcal{D}_I$; e.g. images without birds in them such as an image of a dog or a tree without any bird on it, and poor quality images such as blurry or dark which are difficult to label.
    }
    \label{fig:intro_diag}
\end{figure*}

Even though there is sufficient interest in OOD detection, to the best of our knowledge, it is unclear what precisely entails an OOD input. Existing detectors estimate a distribution that is tied to the training dataset, and flag inputs as OOD when the assigned probability according to the estimated distribution is low. The standard drill involves a set of in-distribution inputs drawn from a dataset such as CIFAR10, and detecting those inputs as OOD that are drawn from a different dataset such as SVHN~\citep{baseline,kaur_udl,mahala}. Such external inputs (from SVHN) would have non-overlapping training class labels (from CIFAR10).
With this in mind, we propose to treat \textit{intended distribution} of images as in-distribution. i.e. images containing semantic information relevant to the training classes irrespective of the background (or spurious) information. \footnote{We will be using the terms ``in-distribution" and ``intended distribution" exchangeably in the paper.}. Inputs deficient of semantic information w.r.t any training class should be detected as OOD. 

Domain generalization or robustness to spurious features in an input is a desired property and a requirement for machine learning models to be put to use~\citep{domain_gen1, domain_gen2}. For instance, as shown in Figure~\ref{fig:intro_diag}, a classifier trained to classify birds in \{sitting birds, flying birds\} is expected to generalize well beyond the training data of birds sitting on trees and birds flying in sky. Inputs from the intended distribution of sitting birds refers to birds sitting on trees, snow or water. Performing OOD detection on inputs with class label in the training classes but outside the training distribution such as birds sitting on snow restricts the utility of the model.

~\cite{spurious_oods} show that the existing detectors are unfortunately tied to the sampling bias of the training dataset. This results in low detection on OOD inputs with spurious features such as background, color, etc. from the training data. The authors report low detection performance of existing detectors on two datasets: 1) Birds~\citep{waterbirds} with class labels in \{waterbirds, landbirds\}, and 2) CelebA~\citep{celeba} with class labels in \{grey hair, non-grey hair\}. Table~\ref{tab:spurious_oods} shows these results for OOD images without birds but containing water (or land) as a spurious feature for waterbirds (or landbirds), and OOD images of bald male with male as a spurious feature for grey hair; examples of these images are shown in Figure~\ref{fig:birds_celeba_imgs}. This means that even though the classifier might be able to generalize better, OOD detectors themselves can stifle its utility. 


\begin{table*}[!t]
\begin{minipage}{0.48\linewidth}
\caption{Low OOD detection by existing detectors on OOD inputs with spurious features from the training data of Birds dataset with class labels in \{waterbirds, landbirds\}, and CelebA dataset with class labels in \{grey hair, non-grey hair\}~\citep{spurious_oods}. Some examples of in-distribution and spurious OOD images for these datasets are shown in Figure~\ref{fig:birds_celeba_imgs}. \textbf{Our Algorithm~\ref{alg:sem_seg_ood_det} significantly improves detection on these OOD inputs deficit of semantic information relevant to any training class: birds for Birds dataset and hair for CelebA dataset.}}
\centering
\small
\begin{adjustbox}{width=1\columnwidth}
\begin{tabular}{c|c|c}
\hline
Detector &  OOD for Birds & OOD for CelebA \\
\hline
Baseline [\citeyear{baseline}] &  25.32 & 16.30  \\
ODIN [\citeyear{odin}] &  22.75 & 18.93  \\
Mahala [\citeyear{mahala}] & 30.65 &  21.25  \\
Energy [\citeyear{energy}] & 25.78 & 28.72  \\
Gram [\citeyear{gram}] & 41.75 & 18.79 \\
Ours & \textbf{98.97} & \textbf{74.36} \\
\hline
\end{tabular}
\end{adjustbox}
\label{tab:spurious_oods}
\end{minipage}\hfill
\begin{minipage}{0.45\linewidth}
    \centering
    \includegraphics[scale=0.25]{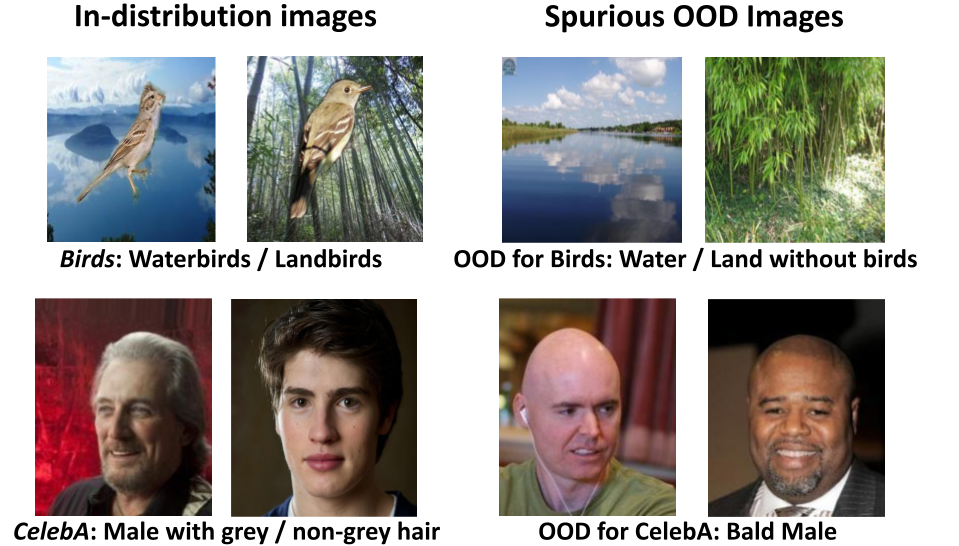}
    \captionof{figure}{Images from Birds and CelebA dataset (left). OOD for birds dataset are images without birds, and OOD for CelebA dataset are images of people without hair (right).}
    \label{fig:birds_celeba_imgs}
\end{minipage}
\end{table*}

The contributions of this paper can be summarized as: \\
\textbf{1. Demystifying OOD Inputs:} Even though there is sufficient interest in OOD detection, to the best of our knowledge, it is unclear what precisely constitutes an OOD input. We propose to model in-distribution for machine learning classifiers as the 
\textit{intended set of images} containing semantic information relevant to the training classes. As a consequence, we define as OOD those inputs whose semantically relevant part is given low probability by the intended distribution.

\textbf{2. OOD Detection based on the Intended Set of Inputs:} We propose two distinct ways of estimating 
the intended set of images for modeling the in-distribution for a classifier. The first one leverages a machine learning model in the presence of a large amount of labeled training data, while the second one utilizes the available expert guidance. We propose two OOD detection algorithms based on the two ways of estimating the intended distribution. 

\textbf{3. Experimental Evaluation:}\\
(a) Table~\ref{tab:spurious_oods} shows that we achieve significant improvement by $57.22\%$ and $45.64\%$ on OOD detection for Birds and CelebA, respectively, with the proposed OOD detection Algorithm~\ref{alg:sem_seg_ood_det} that uses a machine learning model for estimating the indented set. 

(b) Our experiments on COCO~\cite{coco} and MNIST~\cite{lenet} datasets show that the existing detectors overfit to the training data for estimating in-distribution, resulting in (i) false OOD detection on inputs with the same (training class) labels but from a different dataset, and (ii) low OOD detection on inputs whose classes are absent from the set of training classes. This low detection is due to the sensitivity of existing detectors to the spurious features from the training data. The proposed algorithms not only significantly reduce false alarms, but they also improve OOD detection ($\geq 20\%$) on inputs with spurious features from training data.

\textbf{Related Work.} 
OOD detection has been extensively studied and detectors with OOD scores based on the difference in statistical, geometrical or topological properties of in-distribution and OOD inputs have been proposed. These detectors can be classified into three categories, supervised~\citep{mahala,kaur_udl}, self-supervised~\citep{aux,idecode}, and unsupervised~\citep{baseline, odin}. Unsupervised approaches can function without an OOD dataset for training the detector, while supervised approaches do. Self-supervised approaches require a self-labeled dataset for training the detector. This dataset is created by applying transformations to the training data and labeling the transformed data with the applied transformation. The proposed OOD detection algorithms in this paper are unsupervised in nature.~\cite{spurious_oods} show that the existing detectors perform poorly on OOD inputs with spurious features from the training data. They, however, do not propose a solution for fixing the existing detectors.

Domain generalization~\cite{dom_gen_survey} is an active research area where efforts are made for the generalizability of machine learning classifier to its classes beyond the training data. As shown in Figure~\ref{fig:intro_diag}, it tries to ask the question of whether a classifier trained on the images of birds on trees would work on images of birds on water. Domain-invariant representation learning~\citep{dom_inv_learning}, training data augmentation with higher variability~\citep{data_aug} etc. have been proposed to solve this problem. With the intended distribution of images containing (training) class-specific information for a classifier, we propose inputs that do not contain this information as OOD.

There has been a great interest in making use of semantic segmentation networks in scene understanding problem~\citep{sem_seg_adv}, one of the core problems in computer vision with  applications e.g. to autonomous driving, video surveillance, and robot perception~\citep{semantic_seg_imp}. Recently, the use of segmentation networks was proposed to train machine learning classifiers with a handful of training examples~\cite{cvs}. We make use of segmentation networks as the machine learning model for estimating the intended distribution for OOD detection.

\section{Problem Formulation and Methodology}\label{probl_form}
\subsection{Problem Formulation}
Let $(\mathcal{X},\mathcal{A}_\mathcal{X})$ be the measurable space from which images are sampled. We assume that $\mathcal{X}$ is an at most a countable subset of a (possibly very high-dimensional) Euclidean space $\mathbb{R}^{h\times w \times 3}$ whose dimension depends on the size of the images. Here, $h, w$ refer to the height and width of the image, and $3$ stands for the red, green, and blue channels, making the elements of $\mathcal{X}$ colored images.
Let $\Delta(\mathcal{X},\mathcal{A}_\mathcal{X})$ denote the space of probability measures on $(\mathcal{X},\mathcal{A}_\mathcal{X})$, and we consider a candidate distribution  $\mathcal{D}\in\Delta(\mathcal{X},\mathcal{A}_\mathcal{X})$.\footnote{As no confusion arises, we do not distinguish between probability measure and probability distribution.} Let $X_1,\ldots,X_n \sim \mathcal{D}$ be iid, whose realizations $x_1, \dots, x_n$ make the training set $\mathcal{S}$ for a machine learning classifier; $\mathcal{S}:=\{ x_1, \dots, x_n \}$. We assume that the support of $\mathcal{D}$, written $\text{supp}(\mathcal{D})$, is a proper subset of $\mathcal{X}$, that is, $\text{supp}(\mathcal{D}) \subsetneq \mathcal{X}$. 

Now, we introduce what we call the \textit{intended distribution}, i.e. a probability measure $\mathcal{D}_I\in\Delta(\mathcal{X},\mathcal{A}_\mathcal{X})$ whose support $\text{supp}(\mathcal{D}_I)$ is a proper superset of $\text{supp}(\mathcal{D})$. We can write $\text{supp}(\mathcal{D}) \subsetneq \text{supp}(\mathcal{D}_I) \subset \mathcal{X}$. Intended distribution $\mathcal{D}_I$ is needed because it assigns non-zero probability to the set of images which are likely to be seen by the classifier in the real world. For instance, in case of standard birds dataset (Figure~\ref{fig:intro_diag}), the training distribution $\mathcal{D}$ captures images of birds on trees, but the intended distribution $\mathcal{D}_I$ for the classifier can refer to birds on trees,  water, or snow. We define the \textit{intended set of inputs} for a classifier as:
\begin{equation*}
    \mathcal{X}_I := \{ x\in\mathcal{X} : \mathcal{D}_I(x) > \epsilon \} \subset \text{supp}(\mathcal{D}_I),
\end{equation*}
 for some $\epsilon>0$. The measurable space of (class) labels $(\mathcal{Y},\mathcal{A}_\mathcal{Y})$ is assumed to be at most countable.

We ask the following OOD detection question: \textit{Given the training set $\mathcal{S}$ for a machine learning classifier, can we build an OOD detector that is able to detect inputs that lie far from the ones in $\mathcal{X}_I$}?

\subsection{Methodology}
For any image $x\in\mathcal{X}$, we use $\text{rel}(x)$ to denote its semantically relevant part. We propose two approaches to answer the OOD detection question. The first one estimates the intended distribution $\mathcal{D}_I$ with an empirical distribution $\hat{\mathcal{D}}_I$; then, if for a given input $t\in\mathcal{X}$ we have that $\hat{\mathcal{D}}_I(\text{rel}(t))$ is ``too low'', we say that $t$ is OOD. The second one is to build a function that measures the similarity between $\text{rel}(t)$ and the elements of $\mathcal{X}_I$; then, if the similarity is ``too low'', we say that $t$ is OOD. Both these methods aim to detect as OOD an image whose (associated class) label does not belong to $\mathcal{Y}$. 

Algorithm \ref{alg:ood_det} subsumes the two approaches in the generic function $OOD\_Detection_{\hat{\mathcal{X}}_I}$ that depends on the approximation $\hat{\mathcal{X}}_I$ of ${\mathcal{X}}_I$ via the training set $\mathcal{S}$. In the next section, we delve into the details of the two approaches.

\begin{algorithm}[H]
    \caption{Detecting OOD Inputs as Out-of-Intended Distribution Inputs}
    \label{alg:ood_det}
    \begin{algorithmic}
        \STATE {\bfseries Input:} Test datapoint $t \in \mathcal{X}$
        \STATE {\bfseries Parameters:} Training set $\mathcal{S}$, detection threshold $\epsilon$
        \STATE {\bfseries Output:} ``$1$'' if $t$ is detected as OOD; ``$0$'' otherwise
        \STATE $\hat{\mathcal{X}}_I$ = estimated ${\mathcal{X}}_I$ from $\mathcal{S}$ 
        \STATE $OOD\_Detection_{\hat{\mathcal{X}}_I}:\mathcal{X} \rightarrow \mathbb{R}_+$
        \COMMENT{Builds the intended distribution}
        \STATE Return 1 if $OOD\_Detection_{\hat{\mathcal{X}}_I}(t)<\epsilon$, 0 otherwise
    \end{algorithmic}
\end{algorithm}

\vspace{-0.63cm}
\section{Using Semantically Relevant Information for OOD Detection}
\label{sec:technical}
In this section, we  explore the two methods described above. We first elicit our estimation for $\mathcal{X}_I$,
\begin{equation}\label{equation_estim_xi}
    \hat{\mathcal{X}}_I:=\{x^\prime\in\mathbb{R}^{h\times w\times 3}:x^\prime=\mathcal{N}(x), x\in\mathcal{S}\},
\end{equation} 
where $\mathcal{N}$ is a generic segmentation map. $\hat{\mathcal{X}}_I$ is a good estimator of ${\mathcal{X}}_I$ because it preserves the semantically relevant information that is required for classification.

\subsection{Out-of-Intended Distribution Detection with Machine Learning Model}\label{ood_detec_sem_seg}

Let the \textit{oracle classifier} be a map $C:\mathcal{X} \rightarrow \mathcal{Y}$, $\ x \mapsto C(x)=y\in\mathcal{Y}$, which produces the ground truth labels. 
\begin{align*}
    \text{Let } \mathscr{F}:=\{F: \mathcal{X} &\rightarrow \mathcal{X}_I\cup \{\bot\} \text { }|\\
     x&\mapsto F(x)=s \text{, } C(x)=C(s)\},
\end{align*}
where $\bot$ denotes an empty image, and $s$ corresponds to the relevant part $\text{rel}(x)$ of an input image $x$. The elements of $\mathscr{F}$ are the maps that extract the relevant parts of an image while preserving the label assigned by the oracle classifier $C$ to the original image. We require the set $\mathscr{F}$ to satisfy the following two assumptions. Our \textbf{first assumption} is that $\mathscr{F}\neq\emptyset$. This is reasonable since it is almost always the case that we can find a map that extracts the relevant part of an image without losing its label. Now, for any $F\in\mathscr{F}$, let us compute $F(X_1),\ldots,F(X_n)$, where $X_1,\ldots, X_n \sim \mathcal{D}$ (training distribution) are iid. $F(X_1),\ldots,F(X_n)$ are iid random variables distributed according to some distribution $\mathcal{D}^\prime$ on $\mathcal{X}_I \cup \{\bot\}$.\footnote{We tacitly assume that $F$ does not induce correlation; this will always be the case.} Our \textbf{second assumption} is that 
\begin{equation}\label{ass_2}
    d_K(\mathcal{D}^\prime,\mathcal{D}_I) \leq \delta, \quad \text{ for some } \delta \geq 0,
\end{equation}
where $d_K(\mathcal{D}^\prime,\mathcal{D}_I):=\sup_{x \in \mathcal{X}_I}|\mathcal{D}^\prime(x) - {\mathcal{D}}_I(x) |$.\footnote{Here, for simplicity, let $\text{supp}(\mathcal{D}_I)=\mathcal{X}_I\cup \{\bot\}$.} The second assumption is equivalent to saying that we can find a $F\in\mathscr{F}$ such that $F(X_1),\ldots,F(X_n) \sim \mathcal{D}^\prime$ are iid, and (\ref{ass_2}) holds.
This too is reasonable: it states that first sampling an image from $\mathcal{D}$ and then extracting its intended part via $F$ is ``sufficiently similar'' to directly sampling an image from the intended distribution. Then, we have the following.

\begin{theorem}\label{thm_1}
    Let $\mathcal{D}_I$ be defined as above. Then, there exists an estimator $\hat{\mathcal{D}}_I\equiv \hat{\mathcal{D}}_I(n)$ of $\mathcal{D}_I$ depending on the size $n$ of the training set $\mathcal{S}$ such that the following holds almost surely 
    $$\lim_{n\rightarrow\infty} d_K(\hat{\mathcal{D}}_I(n),{\mathcal{D}}_I) \leq \delta.$$
\end{theorem}
We provide the proof of the theorem in supplementary material, where we show that $\hat{\mathcal{D}}_I$ is the empirical measure for $F(X_1),\ldots,F(X_n)$. Theorem \ref{thm_1} states that as the size of the training set increases, the distance between the estimated intended distribution $\hat{\mathcal{D}}_I$ and the true intended distribution $\mathcal{D}_I$ converges to a scalar that is bounded by $\delta$. If the sampling process for the training data was perfect, then first sampling according to $\mathcal{D}$, and then extracting the intended part via $F$ would give exactly the same result as sampling directly from $\mathcal{D}_I$, and $\delta$ would be equal to $0$. The fact that $\delta$ is positive accounts for the error due to the short-fall of the algorithm which estimates the intended distribution from the training data.

In light of Theorem \ref{thm_1}, which shows that -- under two natural assumptions -- the distance between ${\mathcal{D}_I}$ and ${\hat{\mathcal{D}}_I}$ is bounded, we propose to perform OOD detection using $\hat{\mathcal{D}}_I$. In scenarios with a large amount of labeled data available, we propose to use semantic segmentation networks as $\mathcal{N}$ in (\ref{equation_estim_xi}); we denote semantic segmentation networks by $\mathcal{N}_s$ to distinguish them from the expert-guided procedure that we introduce in section \ref{ood_detec_ref_set}. 
The output of a segmentation network $\mathcal{N}_s(x)$, called the \textit{segmentation map}, is the classification of each pixel in the image into either background (semantically irrelevant information) or one of the class labels in $\mathcal{Y}$. We propose $\hat{\mathcal{X}}_I$ as the set of segmentation maps on (the elements of) $\mathcal{S}$, where class information is labeled by the segmentation network, i.e., $\hat{\mathcal{X}}_I:=\{x^\prime:x^\prime=\mathcal{N}_s(x), x\in\mathcal{S}\}$. Here, we call the relevant part of an image as \textit{foreground segment}: the set of pixels in a segmentation map labeled with a class in $\mathcal{Y}$ by $\mathcal{N}_s$.



Segmentation algorithm $\mathcal{N}_s$ filters the input with the class-specific semantic information in $\mathcal{Y}$; since it extracts the relevant part from an input image $x$, we can see $\mathcal{N}_s$ as a map $F\in\mathscr{F}$. Since $\mathcal{N}_s$ extracts class-specific information from an input image without losing its class label \citep{semantic_seg_imp}, we see how the first assumption is satisfied. Then, recall that $X_1,\ldots,X_n \sim\mathcal{D}$ iid, and so $\mathcal{N}_s(X_1),\ldots,\mathcal{N}_s(X_n) \sim\mathcal{D}^\prime$ iid. Distribution $\mathcal{D}^\prime$ satisfies the second assumption as it has been shown that segmentation networks are quite effective at extracting the intended or class-specific foreground data from the training dataset \citep{semantic_seg_imp}.

More rigorously, it is a function $\mathcal{N}_s:  \mathcal{X} \rightarrow \mathbb{R}^{h\times w\times (|\mathcal{Y}|+1)}$. It only keeps the height and width of the image, losing the color information; the third dimension is given by a vector of dimension $|\mathcal{Y}|+1$, where $|\mathcal{Y}|$ is the number of (class) labels, and the extra dimension captures an ``extra label'' associated with the background. 
Its entries are real numbers between $0$ and $1$ that sum up to $1$; they represent the probability of each pixel in an image  belonging to (class) label $y\in\mathcal{Y}$ or to the ``extra label''. 


\textbf{OOD Detection Scores:} 
Classification-based detection scores \citep{baseline,odin} can be put to use in order to perform OOD detection on the foreground segment of an input image. Similar to the baseline detector \citep{baseline} -- which uses the softmax score of the predicted class by a classification network for detection -- we propose to use softmax scores for the predicted class of the foreground segment for detection. Since the detection score must be a single value, we take the average of the softmax scores for the pixels in the foreground segment.  We formalize this score as follows.


Recall that $h$ and $w$ denote the height and width of an image $x\in\mathcal{X}$. Let $H:=\{1,\ldots,h\}$, and $W:=\{1,\ldots,w\}$. For a generic vector $a$, we use $a_i$ to denote its $i$-th entry, while for a generic element $r$ of $\mathbb{R}^{h\times w\times (|\mathcal{Y}|+1)}$, we use $r_{i,j}$ to denote the $(|\mathcal{Y}|+1)$-dimensional vector that we obtain if we ``slice'' $r$ at the first coordinate $i$ and the second coordinate $j$. For $N=|\mathcal{Y}|$, and $q=(q_1,\ldots,q_N,q_{N+1})^\top\in\mathbb{R}^{N+1}$; we define the function $V$ as follows.
\begin{equation*}
 q\mapsto V(q):=\begin{cases}
\max_i q_i  &\text{if } \argmax_i {q}_i \in \{1,\ldots,N\} \\
 0  &\text{otherwise}
\end{cases}.   
\end{equation*}

\begin{definition}\label{bs_def}
  For any $x\in\mathcal{X}$, 
  we define the \textit{baseline score (BLS)} as the average of the softmax scores for pixels in the foreground segment of $x$:
\begin{align}\label{baseline_score}
    BLS(x):=\frac{
    \sum_{i\in H}\sum_{j\in W}V\big(\mathcal{N}_s(x)_{i,j}\big)} {\sum_{i\in H}\sum_{j\in W}\mathbbm{1}\big(V\big(\mathcal{N}_s(x)_{i,j}\big)\neq 0\big)}
\end{align}  

\end{definition}
We can also use the classification-based score used by the ODIN detector \citep{odin}. ODIN is an enhanced version of the baseline detector where the temperature-scaled softmax score of the preprocessed input is used for detection. The input $x$ is preprocessed by adding small perturbations:
$$\widetilde{x} := x - \zeta\text{ sign}(-\nabla_{x} \text{log} \beta^\prime_\star(x,T)).$$ Here 
$\zeta>0$ is the perturbation magnitude,
$\text{sign}$ denotes the sign function,
$T\in\mathbb{R}_{>0}$ is the temperature scaling parameter,
$\beta^\prime(x,T)$ is an $N$-dimensional vector whose $i$-th entry: $\beta^\prime_i(x,T)=\frac{\exp(f_i(x)/T)}{\sum_{j=1}^{N}\exp(f_j(x)/T)}$ is given by the temperature-scaled softmax score of the $i$-th class predicted by the classification network $\mathbf{f} = (f_1, \ldots, f_N)$ that is trained to classify $N$ classes in $\mathcal{Y}$,
and
$\beta^\prime_\star(x,T)=\max_i \beta_{i}^\prime(x,T)$.


\begin{definition}\label{os_def}
  For any $x\in\mathcal{X}$,
  we define the \textit{ODIN score (ODS)} as the average of the softmax scores for pixels in the foreground segment of $\widetilde{x}$:
\begin{align}\label{odin_score}
    ODS(x):=\frac{\sum_{i\in H}\sum_{j\in W}V(\mathcal{N}_s(\widetilde{x})_{i,j})}{\sum_{i\in H}\sum_{j\in W}\mathbbm{1}\big(V\big(\mathcal{N}_s(\widetilde{x})_{i,j}\big)\neq 0\big)}.
\end{align}  
\end{definition}

Then, given an input image $t\in\mathcal{X}$, we can view $ds(t)$, $ds\in\{BLS,ODS\}$, as the estimated intended distribution $\hat{\mathcal{D}}_I$ of Theorem \ref{thm_1} evaluated at the relevant part of $t$, that is, $ds(t)=\hat{\mathcal{D}}_I(\text{rel}(t))$.
We propose Algorithm \ref{alg:sem_seg_ood_det} for OOD detection when $\hat{\mathcal{X}}_I$ is computed according to $\mathcal{N}_s$, and $\mathcal{D}_I$ is estimated by $ds\in\{BLS,ODS\}$.


\begin{algorithm}[H]
    \caption{Out-of-Intended Distribution Detection with Semantic Segmentation Network}
    \label{alg:sem_seg_ood_det}
    \begin{algorithmic}
        \STATE {\bfseries Input:} Test input $t \in \mathcal{X}$, 
        \STATE {\bfseries Parameters:} Semantic segmentation network $\mathcal{N}_s$ trained on $\mathcal{S}$, 
        detection score $ds \in \{BLS, ODS\}$, detection threshold $\epsilon$
        \STATE {\bfseries Output:} ``$1$'' if $t$ is detected as OOD; ``$0$'' otherwise
        \STATE Return 1 if $ds(t) < \epsilon$, 0 otherwise 
    \end{algorithmic}
\end{algorithm}

\subsection{Out-of-Intended Distribution Detection with expert guidance}\label{ood_detec_ref_set}
Datasets such as MNIST, with a history of expert-feature engineering techniques, e.g. shape context \cite{shape_context}, allow semantically relevant pixels to be derived easily. The generation of the segmentation map, which we denote by $\mathcal{N}_r:\mathcal{X} \rightarrow \mathbb{R}^{h \times w \times 3}$ to distinguish it from $\mathcal{N}_s$ in section \ref{ood_detec_sem_seg}, follows a two-step expert-guided process. First, it uses a standard segmentation algorithm to define super (or semantically relevant) pixels of an image. 
Next, it removes the segments which can be regarded as irrelevant (or background) information. This creates an image out of the two components by setting different colors to the semantically relevant and the irrelevant pieces. We leave the details with examples (Fig. 2) to the supplementary material. Following (\ref{equation_estim_xi}), we have $\hat{\mathcal{X}}_I=\{x^\prime\in\mathbb{R}^{h\times w\times 3}:x^\prime=\mathcal{N}_r(x), \ x\in\mathcal{S}\}$. Next, we define a reference set $\mathcal{R} \subset \hat{\mathcal{X}}_I$ as a set of size $|\mathcal{Y}|$ containing one representative of each class $y \in \mathcal{Y}$:
$$\mathcal{R}:=\{x\in\hat{\mathcal{X}}_I : x \sim \text{Unif}(C^{-1}(y)) \text{, } \forall y \in \mathcal{Y}\} \subset \hat{\mathcal{X}}_I.$$
More sophisticated algorithms can be used to replace this simple choice for creating $\mathcal{R}$, such as the ones proposed in ~\cite{interpretable_ood, two_stage_classification}. Nevertheless, we find this simple procedure well-suited for this context.

\textbf{OOD Detection Score:} Here, we use Structural similarity index metric (SSIM)~\cite{ssim} as the OOD detection score. SSIM is a well-known index to compute the statistical similarity between two images. It is calculated as:
\begin{equation}
    \label{eq:ssim_orig}
    \begin{aligned}
    SSIM(x_1, x_2) & = S_1(x_1, x_2) S_2(x_1, x_2), \text{ where} \\ 
    S_1(x_1, x_2) & = lum(x_1, x_2), \text{and} \\
    S_2(x_1, x_2) & = con(x_1, x_2) corr(x_1, x_2)
    \end{aligned}
\end{equation}
The functions $lum$, $con$ and $corr$ compare the luminosity, contrast and correlation between two image inputs $x_1$ and $x_2$. The details of its implementation can be found in \cite{ssim,modified_ssim}, and it permits fast GPU-based implementation.

OOD detection for a test input $t\in\mathcal{X}$ is performed by measuring the SSIM between $\mathcal{N}_r(t)$ and its nearest neighbor in the reference set $\mathcal{R}$. 

\textbf{Algorithm}: Algorithm \ref{alg:normal_ood_det} combines these pieces together. We compute the SSIM of the relevant part $ \text{rel}(t) = \mathcal{N}_r(t)$ of an input image $t \in \mathcal{X}$ with respect to all images in $\mathcal{R}$, and use the maximum value for detection. In other words, if the similarity value of the relevant part $\text{rel}(t)$ of $t$ with its nearest neighbor in $\mathcal{R}$ is below the detection threshold $\epsilon$, we declare $t$ as OOD. 

\begin{algorithm}[H]
    \caption{ Out-of-Intended Distribution Detection with {Reference Set}}
    \label{alg:normal_ood_det}
    \begin{algorithmic}
        \STATE {\bfseries Input:} Test input $t \in \mathcal{X}$
        \STATE {\bfseries Parameters:} Segmentation algorithm $\mathcal{N}_r$, reference set $\mathcal{R}$, detection threshold $\epsilon$
        \STATE {\bfseries Output:} ``$1$'' if $t$ is detected as OOD; ``$0$'' otherwise
        \STATE $v_k=SSIM(r_k,\mathcal{N}_r(t))$, for all $r_k\in\mathcal{R}$
        \STATE Return 1 if $\max_k(v_k) < \epsilon$, 0 otherwise
    \end{algorithmic}
\end{algorithm}
\section{Experiments}
\label{sec:exp}
We perform experiments with the existing state-of-the-art (SOTA) detectors from all the three categories of supervised, unsupervised, and self-supervised OOD detection techniques. 

\textbf{Unsupervised : } Baseline detector~\citep{baseline} is the SOTA unsupervised detector. It uses softmax score of a classifier for the predicted class. ODIN~\citep{odin} is an enhanced version of the baseline detector that uses temperature-scaled softmax score but, for a perturbed input for detection. Details are in section~\ref{ood_detec_sem_seg}.

\textbf{Supervised : } Mahalanobis detector (Mahala) is the SOTA supervised detector which uses Mahalanobis distance~\citep{mahalanobis-distance} of the input in the training
feature space of the classifier for detection. 

\textbf{Self-supervised :} Aux~\citep{aux} is the SOTA self-supervised detector which uses error in the prediction of the applied transformation on the input for detection. 
It trains a classifier with an auxiliary task of predicting the applied rotation, vertical and horizontal translations. The sum of the error in the three predictions and classification error is used for detection.


\textbf{Evaluation Metrics:} We call in-distribution inputs as positives and OOD inputs as negatives. We report the Receiver Operating Characteristic curve (ROC), Area under ROC (AUROC), and True Negative Rate (TNR) at $95\%$ True Positive Rate (TPR) for evaluation. These are the standard metrics used in OOD detection~\citep{aux, odin, mahala}.

\subsection{Case Study I: OOD Detection with Semantic Segmentation Network}
\subsubsection{\textbf{Dataset and Motivation}} Common Objects in Context-Stuff (COCO) dataset~\citep{coco_stuff} is a large-scale vision dataset created for the purpose of training machine learning models for object detection, segmentation and captioning with $182$ object classes. We use that subset (training and test) of COCO  which can be classified with the class labels from the set $\mathcal{Y} = $ \{cup, umbrella, orange, toaster, broccoli, banana, vase, zebra, kite\}. These classes share the same label space with another dataset Vizwiz~\citep{vizwiz}. Vizwiz is a real patient dataset captured by visually impaired people, with the purpose developing algorithms for assistive technologies. Where the quality of images captured can be an issue. So, images in the Vizwiz are labeled with either ``no issues'', or with issues such as  ``blurry'', ``too bright'', ``too dark'', ``camera obstructed'' etc. We call the images with ``no issues'' label in the Vizwiz dataset as the \textit{clear Vizwiz}.  

We train the ResNet18~\citep{resnet} model to classify the training set of COCO dataset.  With $74.33\%$ as model's accuracy on test COCO, it achieves a comparable accuracy of $68.14\%$ on the clear Vizwiz. Detecting inputs from clear Vizwiz as OOD by the existing detectors restricts the generalizability of classifiers from the training distribution $\mathcal{D}$ to the intended distribution $\mathcal{D}_I$. 



\begin{figure*}[!t]
        \centering
        \includegraphics[width=2\columnwidth]{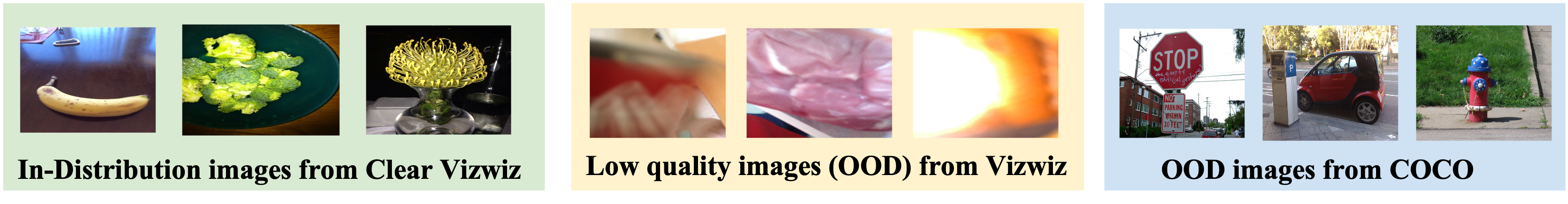}
    \caption{Examples of images from the three test cases for COCO.}
    \label{fig:coco_test_cases}
\end{figure*}


\subsubsection{\textbf{Semantic Segmentation Network $\mathcal{N}_s$ for Algorithm~\ref{alg:sem_seg_ood_det} and Classifier for Detection by Existing Detectors}} 
\label{sec:trained_sem_seg_network}
As recommended by the authors of the COCO dataset~\citep{coco_github}, we train the DeepLab version 2 (v2) segmentation network ~\citep{deeplab} on the training set of COCO. 
DeepLab v2 uses ResNet101~\citep{resnet} as the backbone model. For a fair comparison with the existing detectors, we train the ResNet101 classifier on the training set of COCO. We use the trained classifier for OOD detection by the existing SOTA unsupervised and supervised detectors. The accuracy of the classifier on the test COCO set is $68.64\%$. COCO dataset is commonly used for object detection and segmentation. The classification accuracy of 68.64\% is comparable with the SOTA detection accuracy (in terms of mean average precision) of $64.2\%$ on COCO~\citep{sota_coco_det_accuracy}. For the self-supervised detector AUX, we train the ResNet101 classifier with the auxiliary losses of rotations and translations. Its classification accuracy on the test COCO set is $74.11\%$.


\subsubsection{\textbf{Test Cases and Results}}
We consider the following three test cases:
\\ \textbf{(a) In-Distribution from Clear Vizwiz:} Inputs with the class labels in $\mathcal{Y}$ but from clear Vizwiz. 
\\ \textbf{(b) OOD from Vizwiz:} Inputs with blurry, too bright, too dark, and obstructed issues from Vizwiz. Due to the quality issues of these images, this dataset cannot be labeled with any labels in $\mathcal{Y}$.
\\ \textbf{(c) OOD from COCO:} Inputs from test COCO dataset with class labels not in $\mathcal{Y}$. Here, we filter that subset of test COCO that can be classified with class labels from the set \{traffic light, stop sign, parking meter, fire hydrant\}.

\begin{figure*}[!t]
\begin{subfigure}
        \centering
        \includegraphics[width=1\columnwidth]{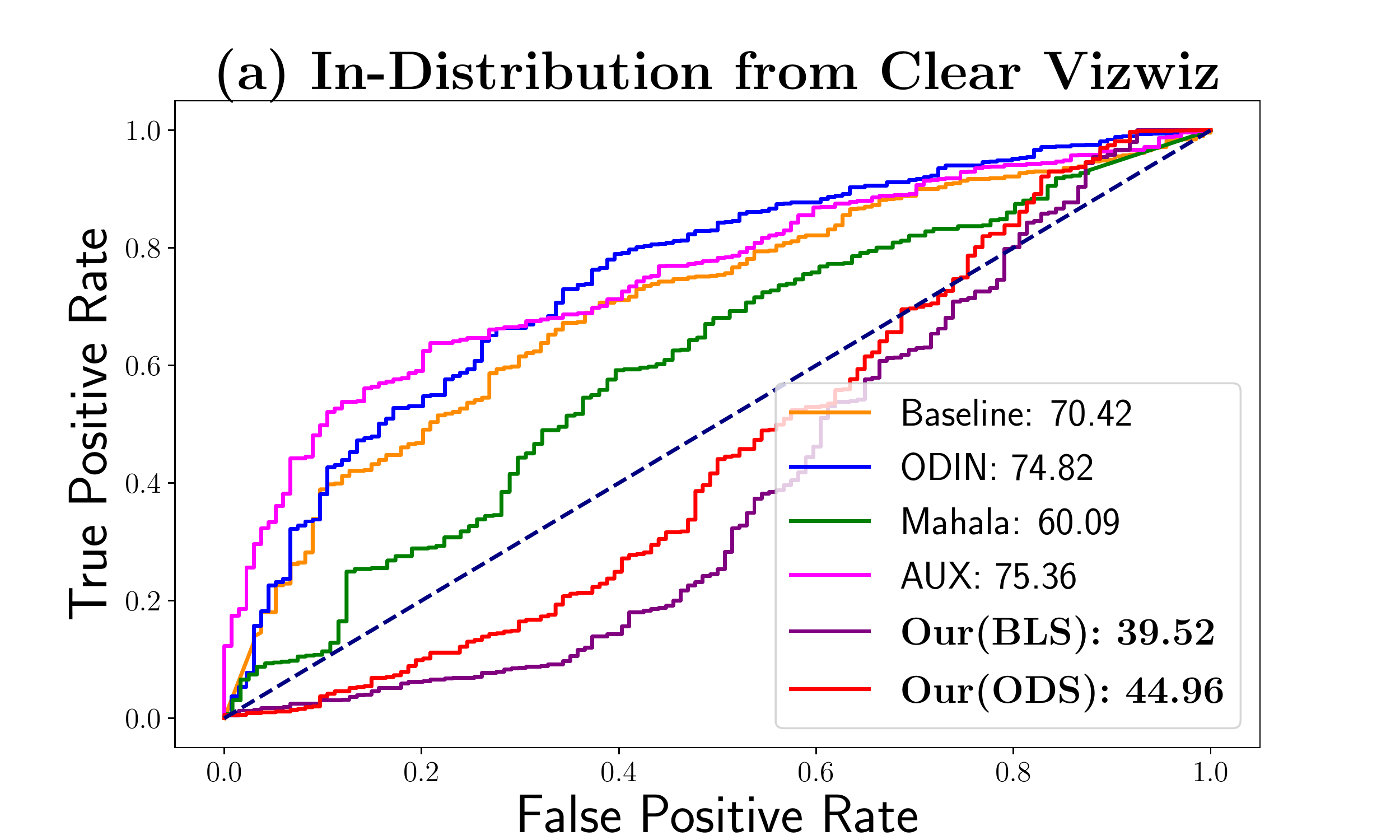}
    \end{subfigure}
    \begin{subfigure}
        \centering
        \includegraphics[width=1\columnwidth]{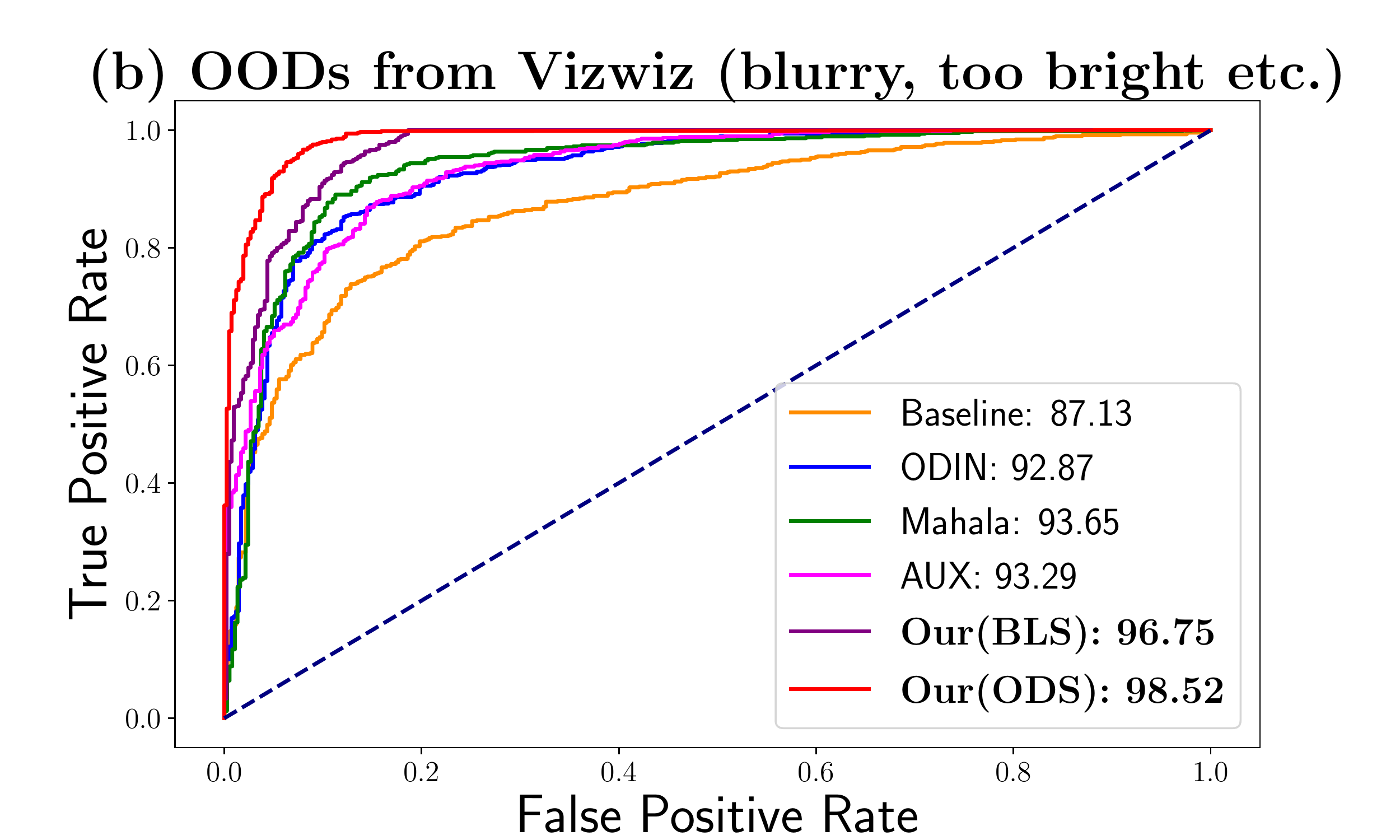}
    \end{subfigure}
    \begin{subfigure}
        \centering
        \hspace*{4cm}
        \includegraphics[width=1\columnwidth]{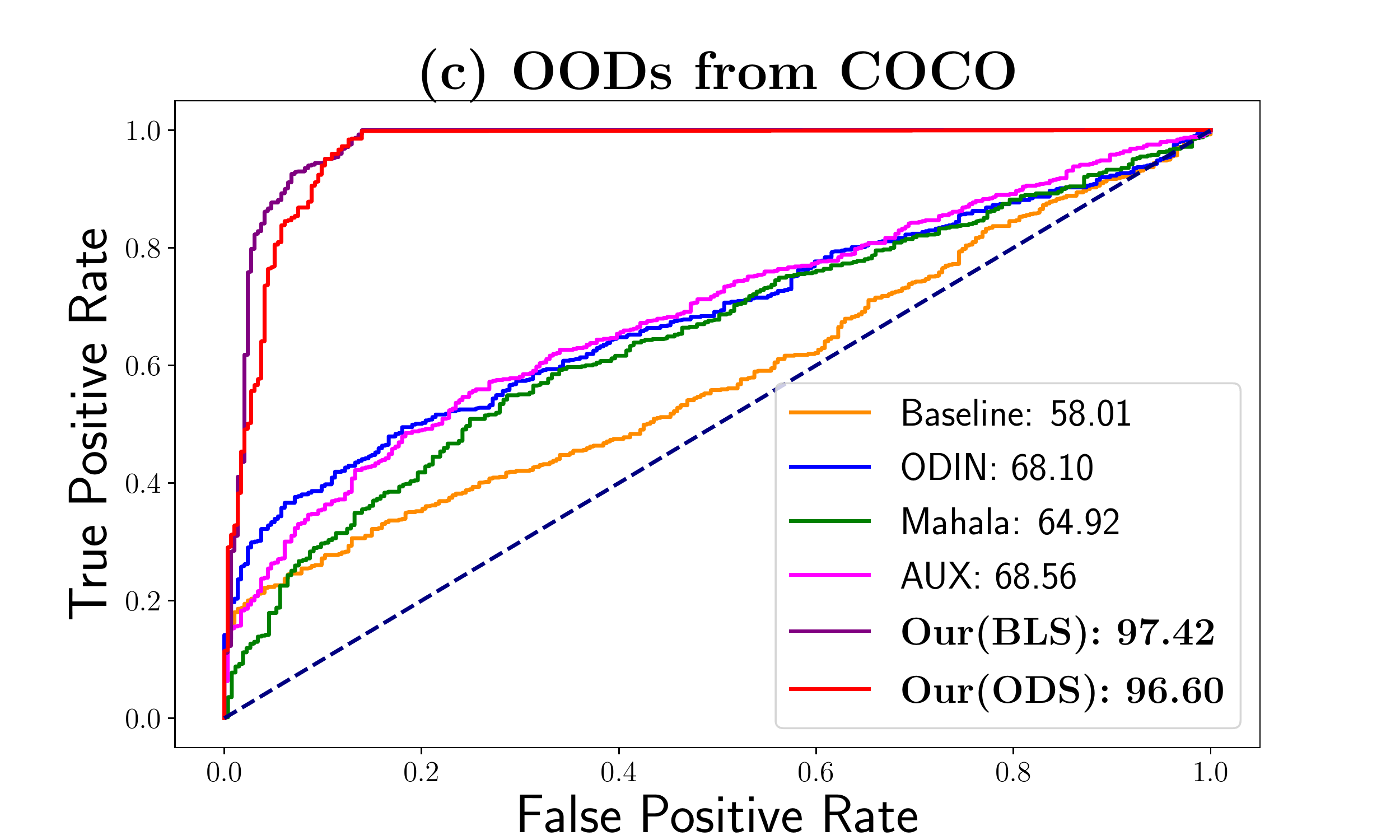}
    \end{subfigure}
\caption{
AUROC less than $50\%$ on in-distribution inputs from clear Vizwiz (a), and highest AUROC on OOD inputs from Vizwiz (b) as well as on OOD inputs from COCO (c) by our Algorithm~\ref{alg:sem_seg_ood_det} shows that the proposed detection not only significantly reduces false alarms but also improves on OOD detection on inputs with spurious features from the training set.}
\label{fig:seg_coco_b_c_results}
\end{figure*}

Figure~\ref{fig:coco_test_cases} shows some examples of the images from the three test cases. 
Figure~\ref{fig:seg_coco_b_c_results} compares the ROC and AUROC results of the existing detectors with Algorithm~\ref{alg:sem_seg_ood_det} on these cases: 
\\ \textbf{(a) In-Distribution from Clear Vizwiz (Fig.~\ref{fig:seg_coco_b_c_results}(a))}: AUROC less than $50\%$ by our approach (with both the baseline and ODIN scores) implies that the proposed detector is not able to distinguish between the test COCO and Clear Vizwiz datasets. AUROC greater than $50\%$ by the existing detectors implies that these detectors distinguish clear Vizwiz from the test COCO by assigning higher OOD detection scores to clear Vizwiz.
\\ \textbf{(b) OOD from Vizwiz (Fig.~\ref{fig:seg_coco_b_c_results}(b))}: With these images as OOD for COCO, we require the AUROC to be as close to one as possible. We achieve the best AUROC of $98.52$ with ODS and the second best AUROC of $96.75$ with BLS.
\\ \textbf{(c) OOD from COCO (Fig.~\ref{fig:seg_coco_b_c_results}(c))}: Significantly higher ($\geq 28.86\%$) AUROC by Algorithm~\ref{alg:sem_seg_ood_det} (with both scores) than the existing ones indicates that the proposed detector performs OOD detection on these inputs with spurious features from the training data better than the existing ones. 

We also perform additional experiments on existing benchmarks where OOD datasets such as SVHN~\citep{svhn}, Imagenet~\citep{imagenet}, and LSUN~\citep{lsun} are considered. These results are included in the supplementary material. Here also, we perform the best (in terms of AUROC) for detection on OOD inputs from Imagenet and LSUN. For SVHN, Mahala (supervised detector) performs the best  with $99.04\%$ AUROC, and the result of Algorithm~\ref{alg:sem_seg_ood_det} (unsupervised detection) is $97.25\%$ AUROC.

\subsection{Case Study II: OOD Detection with a Reference Set}
\subsubsection{\textbf{Dataset and Motivation}}
We use a mixture of MNIST-M~\citep{mnist-m} and Background-Colored-MNIST (BC-MNIST)~\citep{background_colored_mnist} datasets. Both MNIST-M and BC-MNIST are modified versions of MNIST~\citep{lenet} dataset. MNIST-M is MNIST with its digits blended over patches of colored images. BC-MNIST is the colored version of MNIST where both digits and background are colored.  We use a mixture dataset of $100\%$ data from MNIST-M and $50\%$ data from BC-MNIST. We call this dataset as \textit{Mix-MNIST.}

With $60,000$ training images in MNIST-M and $4000$ training images in BC-MNIST, $96.77\%$ of the training data in Mix-MNIST comes from MNIST-M and the remaining $3.23\%$ from BC-MNIST. We train the LeNet5~\citep{lenet} classifier on Mix-MNIST.
The classifier achieves comparable accuracy of $90\%$ and $91\%$ on test MNIST-M and test BC-MNIST datasets respectively. Therefore, with the classifier's ability to generalize on BC-MNIST with only $3.23\%$ of BC-MNIST as the training data, detecting inputs from BC-MNIST as OOD by the exiting detectors (Fig.~\ref{fig:mnist_auroc_tnr}) limits the applicability of the classifier. 




\subsubsection{\textbf{Experimental Details and Results}}
\label{sec:trained_seg_algo}
\begin{figure}[!t]
        \centering
        \includegraphics[width=1\columnwidth]{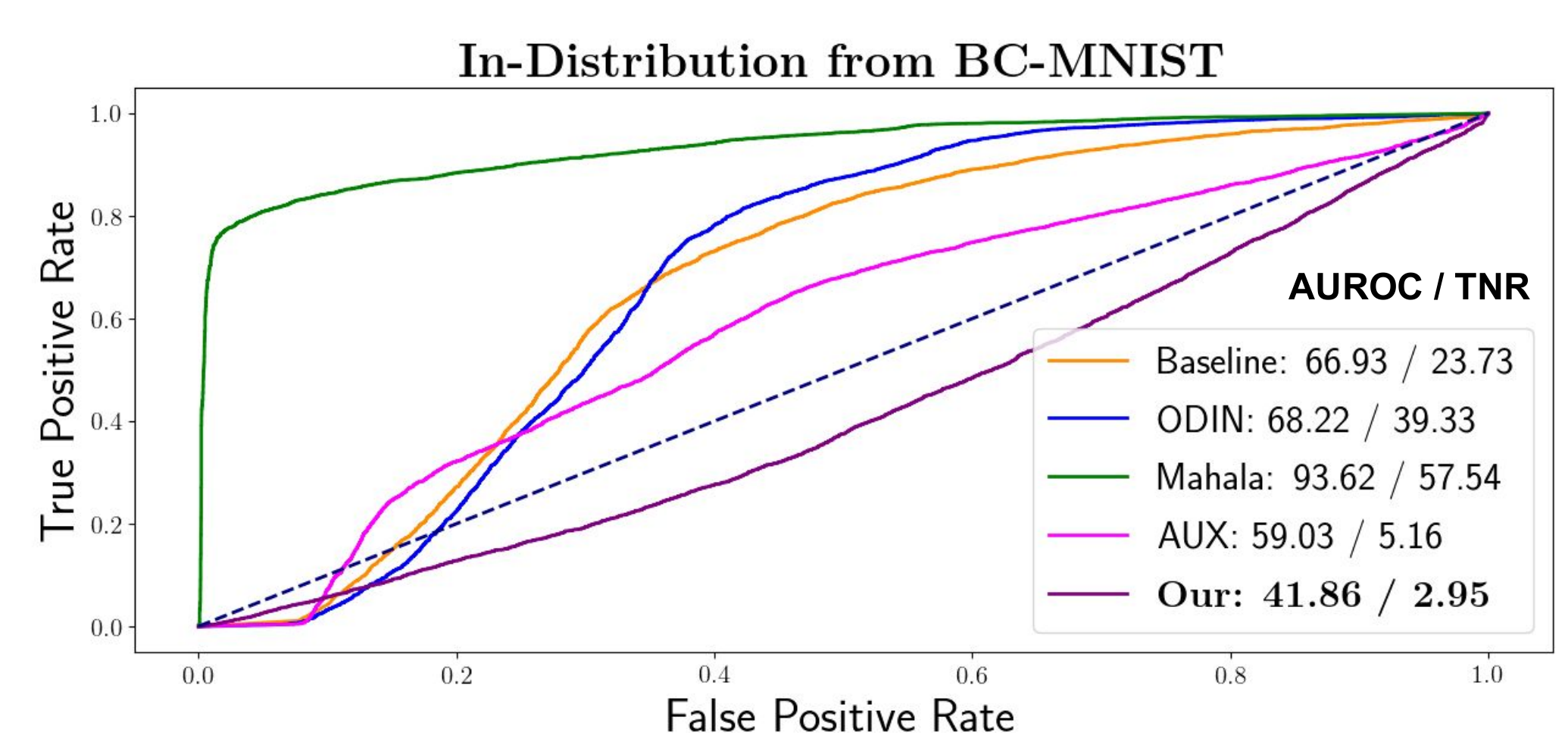}
\caption{AUROC less than $50\%$ on in-distribution inputs from BC-MNIST by Algorithm~\ref{alg:normal_ood_det} shows that the proposed algorithm significantly reduces false alarms (by the existing detectors) for the training set of Mix-MNIST.}
\label{fig:mnist_auroc_tnr}
\end{figure}

For the existing detectors, we use trained the LeNet5 model with its accuracy of $91.91\%$ on the test set of Mix-MNIST. 
Figure~\ref{fig:mnist_auroc_tnr} compares the ROC, AUROC, and TNR results of the existing detectors with Algorithm~\ref{alg:normal_ood_det} on the test set of BC-MNIST. 
AUROC  higher than $50\%$ by the existing detectors implies that existing detectors distinguish the test data of Mix-MNIST from the test set of BC-MNIST with higher OOD detection scores assigned to BC-MNIST. AUROC less than $50\%$ by the proposed Algorithm~\ref{alg:normal_ood_det} shows that it does not distinguish between the test sets of Mix-MNIST and BC-MNIST. We also achieve the lowest false alarm rate of $2.95\%$ here. 
We perform additional experiments for Mix-MNIST with OOD datasets from (low quality) Vizwiz and Fashion-MNIST. Details and results on these experiments are included in the supplementary material. 



\section{Conclusion}
\label{sec:conc}
In this paper, we make use of the training class-specific semantic information for explicitly defining and detecting OOD inputs to a machine learning classifier. We show that including more nuanced semantic information about the content of images can improve OOD detection significantly. This, to the best of our knowledge, is one of the first approaches which differentiates between training distribution and intended distribution.
\section{Appendix}
\label{sec:appendix}
\subsection{Proof of Theorem~\ref{thm_1}}
\begin{proof}
Call $\mathcal{D}^\prime_n$ the empirical measure for $F(X_1),\ldots,F(X_n)$. We have that
\begin{align}
    \lim_{n\rightarrow\infty} d_K(\mathcal{D}^\prime_n,{\mathcal{D}}_I) &\leq \lim_{n\rightarrow\infty} \left[ d_K(\mathcal{D}^\prime_n,\mathcal{D}^\prime) + d_K(\mathcal{D}^\prime,{\mathcal{D}}_I) \right] \label{first_ineq}\\
    &= \lim_{n\rightarrow\infty}  d_K(\mathcal{D}^\prime_n,\mathcal{D}^\prime) + \lim_{n\rightarrow\infty} d_K(\mathcal{D}^\prime,{\mathcal{D}}_I) \label{first_eq}\\
    &= \lim_{n\rightarrow\infty}  d_K(\mathcal{D}^\prime_n,\mathcal{D}^\prime) + d_K(\mathcal{D}^\prime,{\mathcal{D}}_I) \nonumber\\
    &\leq 0+\delta = \delta \label{second_ineq}
\end{align}
almost surely, where (\ref{first_ineq}) comes from the triangular inequality, (\ref{first_eq}) comes from the linearity of the limit operator, and (\ref{second_ineq}) holds because $\lim_{n\rightarrow\infty}  d_K(\mathcal{D}^\prime_n,\mathcal{D}^\prime)=0$ almost surely by Glivenko-Cantelli's Theorem, and $d_K(\mathcal{D}^\prime,{\mathcal{D}}_I)\leq \delta$ by our second assumption. The proof is completed by putting $\hat{\mathcal{D}}_I(n)=\mathcal{D}^\prime_n$.
\end{proof}

\subsection{Example Images from COCO segmented with class-specific relevant information}
\begin{figure}[!h]
    \centering
    \includegraphics[scale=0.5]{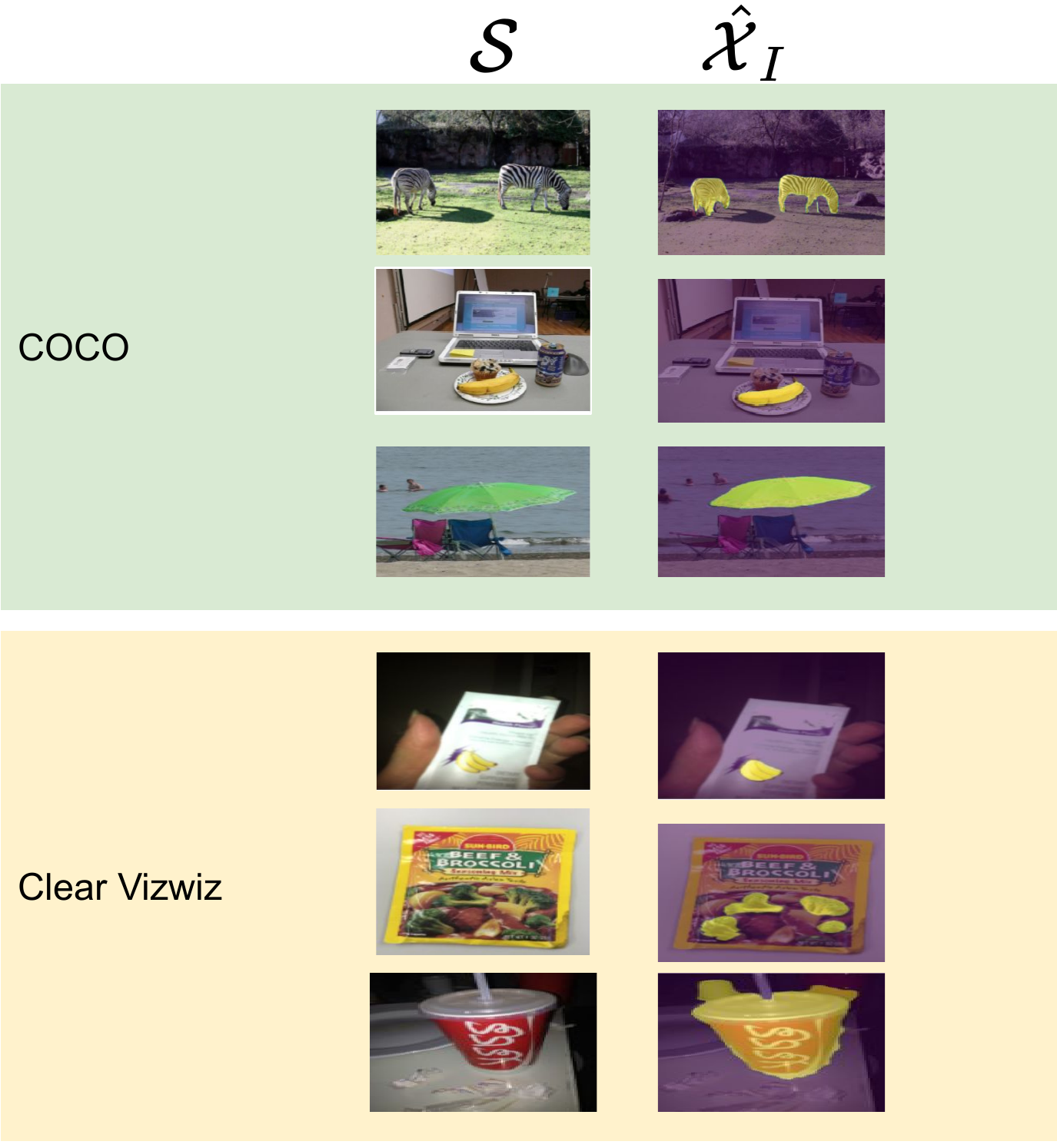}
    \caption{Example images from COCO and clear Vizwiz on the left and output of the trained semantic segmentation network $\mathcal{N}_s$ on these images on the right. Yellow color in the segmented images represents the class-specific semantically relevant information, and the purple color represents the semantically irrelevant or background information.}
    \label{fig:coco_sem_seg}
\end{figure}

Figure~\ref{fig:coco_sem_seg} shows some examples of images sampled from COCO and clear Vizwiz datasets on the left and  corresponding output of the trained semantic segmentation network $\mathcal{N}_s$ from section~\ref{sec:trained_sem_seg_network} on the right. 




\subsection{Additional Results on Existing benchmarks}
\begin{table*}[!h]
\centering
\begin{tabular}{c|ccccc}
\hline
OOD       
        & Baseline & ODIN  & Mahala & Aux & Ours (BLS)                                                                                             \\ 
\hline
SVHN & 89.08  & 94.23 & \textbf{99.04} & 98.79 & 97.25 \\ 
Imagenet &  86.02 & 91.59 & 95.68 & 95.31 & \textbf{97.31} \\
LSUN & 83.24 & 91.24 & 94.28 & 96.21 & \textbf{99.57} \\
\bottomrule
\end{tabular}
\caption {\label{tab:existing_benchmarks}
Comparison of Algorithm~\ref{alg:sem_seg_ood_det} with SOTA detectors on the existing benchmarks.} 
\end{table*}

\subsection{Details about the two-step Segmentation Algorithm $\mathcal{N}_r$ used in Algorithm~\ref{alg:normal_ood_det}}
\label{sec:two_step_algo}
\begin{figure}[!h]
    \centering
    \includegraphics[scale=0.5]{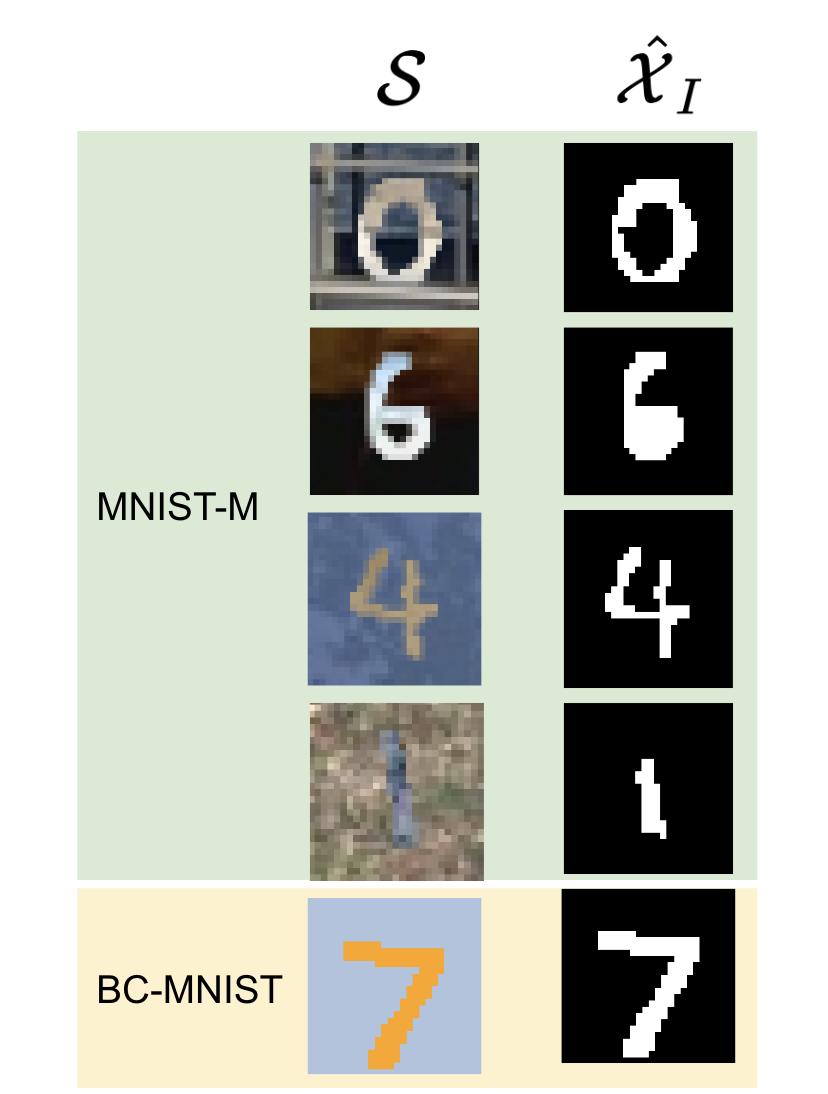}
    \caption{Example images from Mix-MNIST on the left and the output of the segmentation algorithm $\mathcal{N}_r$ on these images on the right. White color in the segmented images represents the class-specific semantically relevant information, and the black color represents the semantically irrelevant information.}
    \label{fig:seg_mnist}
\end{figure}

Detecting semantically relevant pixels is the first step in this algorithm.
In order to separate the semantically relevant pixels, we first partition the image into meaningful segments  using Felzenszwalb's Algorithm \cite{felzenszwalb2004efficient}. Next we mark the segments placed away from the center as being semantically irrelevant. Whatever remains closely maps to semantically relevant information. We binarize the result in the previous step, to obtain a black and white version of the image.
Figure~\ref{fig:seg_mnist} shows some examples of images sampled from Mix-MNIST on the left and  corresponding output of the  segmentation algorithm $\mathcal{N}_r$ from section~\ref{sec:trained_seg_algo} on the right.

\subsection{Experimental Details and Additional Experiments on Mix-MNIST}

\subsubsection{Experimental Details}
\begin{figure*}[!h]
    \centering
    \includegraphics[scale=0.5]{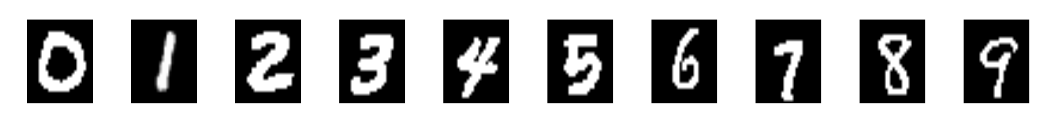}
    \caption{Reference set of 0-9 digits extracted from images in Mix-MNIST by the segmentation algorithm $\mathcal{N}_r$.}
    \label{fig:ref_set}
\end{figure*}

Given two binarized versions of an image pair by the segmentation algorithm $\mathcal{N}_r$ described in Appendix~\ref{sec:two_step_algo}, we compute the SSIM value between these images. We restrict ourselves to a non-negative version of the SSIM metric in this paper. To estimate whether an image contains digit, we maintain a reference set for digits zero to nine. Figure~\ref{fig:ref_set} shows the reference set used in experiments. For a given test image, we compute the SSIM between the binary version of the image and each digit image in the reference set. If the test image does not resemble any digit in the reference set, we declare it to be OOD.

\subsubsection{Additional Experiments}
We conduct additional experiments on Mix-MNIST with the following two test cases:
\\
\textbf{(a) OOD from Vizwiz:} Images with the blurry, too dark, and obstructed quality issues from Vizwiz.
\\\textbf{(b) OOD from Fashion-MNIST:} Images from Fashion-MNIST~\citep{fashion-mnist} dataset with class labels from fashion objects such as trousers, shoe etc.

\begin{figure*}[!h]
    \begin{subfigure}
        \centering
        \includegraphics[width=0.5\linewidth]{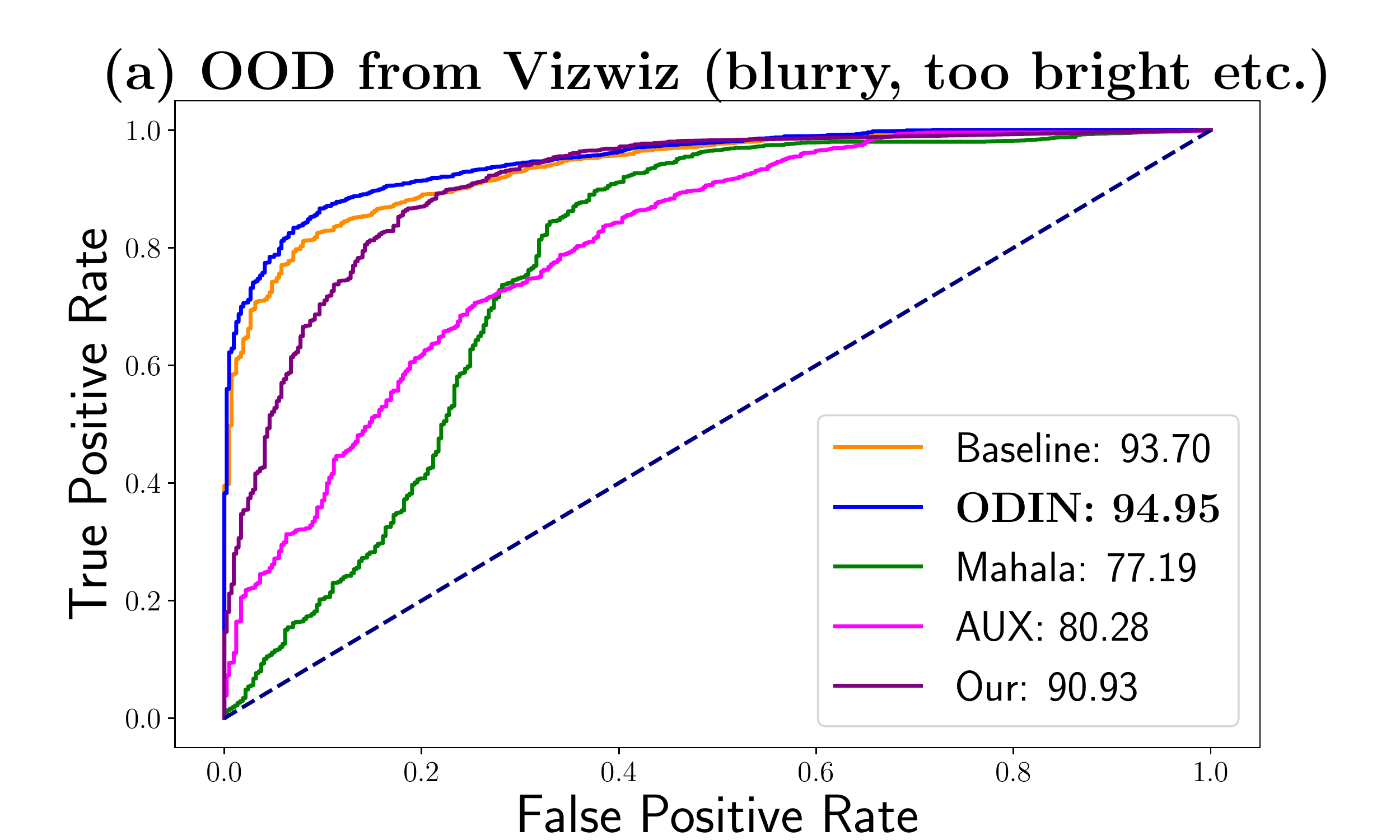}
    \end{subfigure}
    \quad
    \begin{subfigure}
        \centering
        \includegraphics[width=0.5\linewidth]{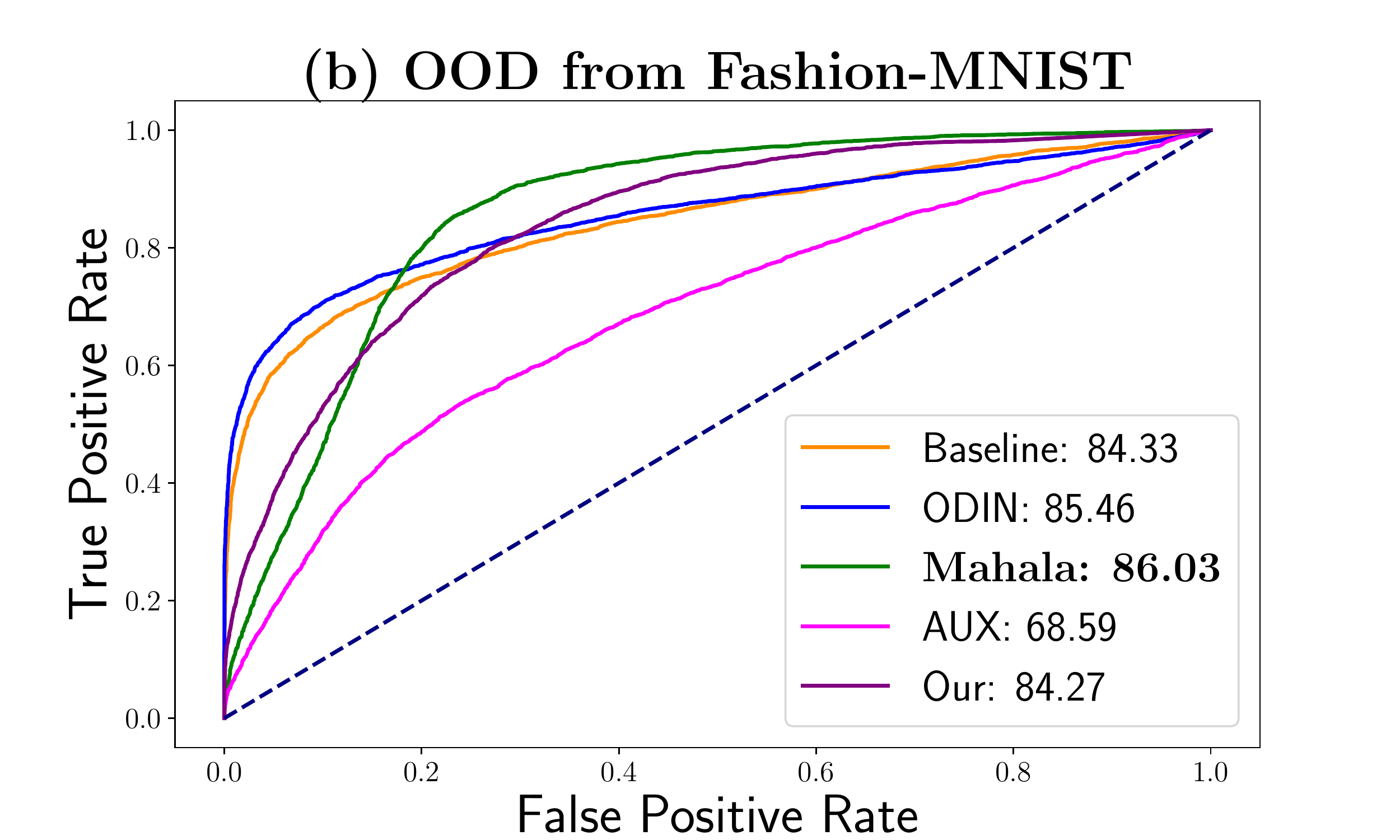}
    \end{subfigure}
\caption{AUROC results of the existing detectors and Algorithm~\ref{alg:normal_ood_det} on Mix-MNIST.}
    \label{fig:mnist_auroc}
\end{figure*}

The results are as follows:
\\
Figure~\ref{fig:mnist_auroc} compares the ROC and AUROC results of the existing detectors with the proposed OOD detection Algorithm~\ref{alg:normal_ood_det}. Table~\ref{tab:mnist_tnr} shows these results on TNR (at 95\% TPR) on these test cases:
\\\textbf{(a) OOD from Vizwiz (Fig.~\ref{fig:mnist_auroc}(a))}: With failure to assign any labels to this dataset due to quality issues, these images are OOD for the Mix-MNIST dataset and here we require the AUROC to be as close to one as possible. The existing detector ODIN achieves the best AUROC of $94.95\%$ and our result is $90.93\%$. We achieve the best TNR ($@ 95\% $TPR) detection of $67.15\%$ here. 
\\ \textbf{(b) OOD from Fashion-MNIST (Fig.~\ref{fig:mnist_auroc}(b))}: With the class labels of Fashion-MNIST disjoint from  the classes in Mix-MNIST, images from  Fashion-MNIST are OOD for Mix-MNIST. The existing supervised detector Mahala achieves the best AUROC of $86.03\%$ and our (unsupervised) results are comparable at $84.27\%$. Mahala achieves the best TNR ($@ 95\% $TPR) detection of $56.86\%$ and ours is second best at $44.67\%$.

\begin{table*}[!h]
\caption{TNR (@95\% TPR) of existing detectors and Algorithm~\ref{alg:normal_ood_det} on Mix-MNIST.}
\centering
\small
\begin{tabular}{c|c|c|c|c|c}
\toprule
Test Set &  Baseline  & ODIN & Mahala &  AUX & Our \\
\hline
OOD from Vizwiz \ \ \ & 64.73 & 66.67 & 54.16 & 43.23 & \textbf{67.15} \\
OOD from Fashion-MNIST \ \ \  & 23.13 & 19.02 & \textbf{56.86} & 11.24 & 44.67  \\
\bottomrule
\end{tabular}
\label{tab:mnist_tnr}
\end{table*}

\subsection{Details about the experiments on Birds and CelebA dataset}
We compare TNR (at $95\%$ TPR) for existing detectors on Birds and CelebA datasets for OOD detection on Spurious OOD test set, as reported by~\citep{spurious_oods}.
\subsubsection{CelebA}
We use the semantic segmentation network $\mathcal{N}_s$ by~\citet{face_parsing} on CelebA dataset. The network segments faces into different parts including nose, hair, mouth, etc. In this experiment, we ran Algorithm~\ref{alg:sem_seg_ood_det} with the Baseline score. 

\subsubsection{Birds}
We use the semantic segmentation network $\mathcal{N}_s$ by \citet{segmentation_backbone}. Here, we select Feature Pyramid Network (FPN)~\citep{FPN} with ResNet50~\citep{resnet} as its backbone architecture. It segments the images into two parts: bird and background. In this experiment, we ran Algorithm~\ref{alg:sem_seg_ood_det} with the Baseline score. 



\newpage
\bibliography{main}
\end{document}